\documentclass{article}
\usepackage{spconf,amsmath,graphicx}

\usepackage{amsfonts} 
\usepackage[colorlinks,linkcolor=blue]{hyperref}
\usepackage{mwe}

\title{Self-semantic contour adaptation for cross modality \\brain tumor segmentation}
\name{Xiaofeng Liu, Fangxu Xing, Georges El Fakhri, Jonghye Woo}
\address{Dept. of Radiology, Massachusetts General Hospital and Harvard Medical School, Boston, MA, USA
}

\begin{document}
%
\maketitle
\begin{abstract}
Unsupervised domain adaptation (UDA) between two significantly disparate domains to learn high-level semantic alignment is a crucial yet challenging task.~To this end, in this work, we propose exploiting low-level edge information to facilitate the adaptation as a precursor task, which has a small cross-domain gap, compared with semantic segmentation.~The precise contour then provides spatial information to guide the semantic adaptation. More specifically, we propose a multi-task framework to learn a contouring adaptation network along with a semantic segmentation adaptation network, which takes both magnetic resonance imaging (MRI) slice and its initial edge map as input.~These two networks are jointly trained with source domain labels, and the feature and edge map level adversarial learning is carried out for cross-domain alignment. In addition, self-entropy minimization is incorporated to further enhance segmentation performance. We evaluated our framework on the BraTS2018 database for cross-modality segmentation of brain tumors, showing the validity and superiority of our approach, compared with competing methods.

\end{abstract}
\section{Introduction} \vspace{-5pt} 

Accurate tumor segmentation plays a vital role in the early diagnosis and surgical planning. Over the past several years, due to advances in deep learning, segmentation performance has been substantially improved, when compared with prior approaches. In particular, deep learning-based semantic segmentation of brain tumors using magnetic resonance imaging (MRI) data has been actively developed, the goal of which is to make pixel-wise semantic classification \cite{yuheng2017image,rampun2019breast,liu2021adapting}. For example, each voxel seen in the MRI volumes of the brain can be categorized as the enhancing tumor (EnhT), peritumoral edema (ED), necrotic and non-enhancing tumor core (CoreT), or healthy tissue \cite{menze2014multimodal}. To date, many deep learning based approaches have relied on the independent and identically distributed (i.i.d.)~assumption between training and testing data~\cite{liu2021adversarial}. Yet, this assumption is often violated, when multiple acquisition parameters or imaging modalities are involved, resulting in a significant performance degradation \cite{yuheng2017image,rampun2019breast,liu2021adapting}.



Cross-modality unsupervised domain adaptation (UDA), therefore, has been proposed to mitigate this performance degradation, when applying a network trained on one MRI modality (e.g., T2-weighted MRI) to another (e.g., T1-weighted MRI) \cite{zou2020unsupervised,han2021deep}. Cross-modality UDA aims to transfer knowledge learned from a labeled source modality to an unlabeled target modality to deal with the difficulty of labeling in the target modality \cite{liu2021subtype,liu2021generative,liu2021domain,liu2021recursively}. ~Adversarial training approaches~\cite{liu2021mutual} would be a typical solution to this problem, by enforcing features extracted from source and target domains to be indistinguishable.~Simply aligning the large semantic feature discrepancy, however, may result in negative transfer, thereby degenerating the performance in both domains \cite{liu2021adversarial}.~As such, accurate alignment of a large domain gap in a cross-modality setting remains a long-lasting challenge \cite{han2021deep}.

\begin{figure*}[t]
\begin{center}
\includegraphics[width=1\linewidth]{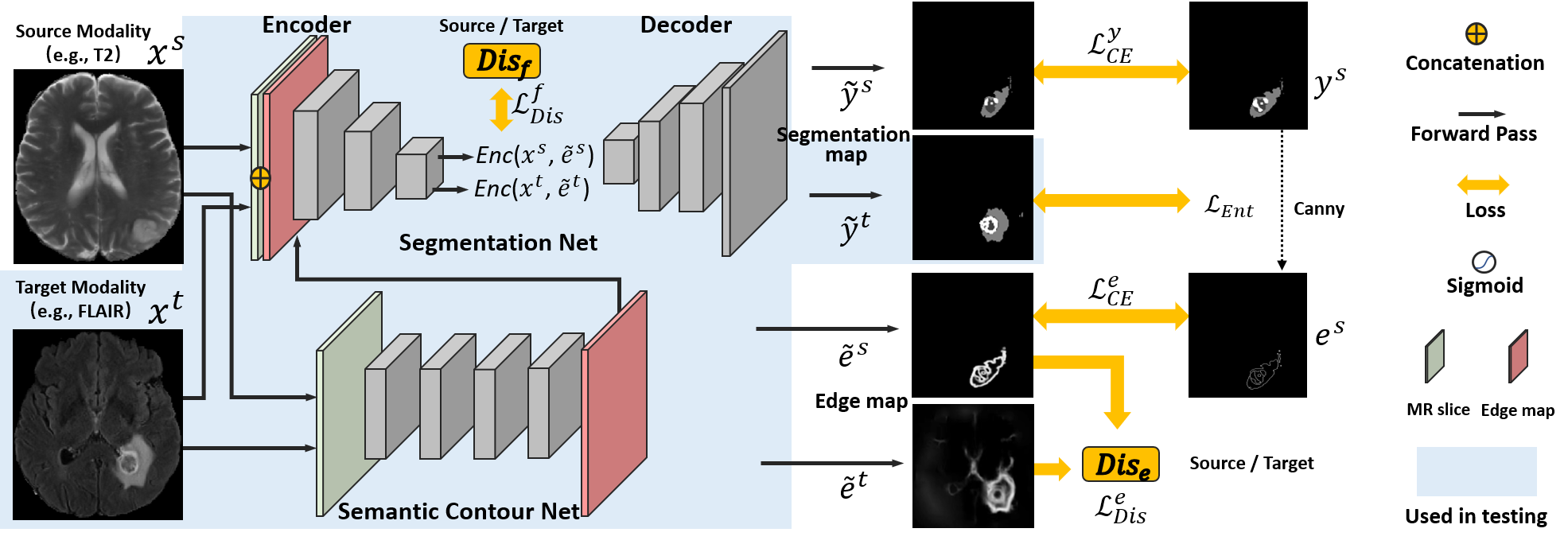} \vspace{-30pt} 
\end{center} 
\caption{Illustration of our proposed multi-task adaptation framework, comprising a consecutive semantic edge delineation and an edge conditioned segmentation module. The edge map and feature level discriminators are applied to edge and segmentation adaptation, respectively. We note that only the light blue masked modules are used in testing.} 
\label{fig:illus}\vspace{-5pt} 
\end{figure*} 

To tackle the aforementioned issues, in this work, we propose to leverage edge delineation that is closely related to semantic segmentation, while being relatively ``easier" to adapt \cite{yuheng2017image}. Actually, semantic segmentation aims to achieve not only an edge delineation, but also a semantic categorical label assignment \cite{yuheng2017image,rampun2019breast}. In addition, there is considerable evidence supporting that the low-level edge features in the shallow layers of convolutional neural networks (CNN) are highly transferable, and have a smaller domain gap than semantic segmentation adaptation \cite{long2015learning}. Accordingly, the adaptation for the edge delineation can serve as a precursor task, and provide relatively precise contours for target domain samples. Since there is a strong correlation between the semantic edge delineation and semantic segmentation in that the edges are the boundaries of the semantic classes, we are able to facilitate the subsequent segmentation with the edge map. We note that we are only interested in the semantic contours of the to-be segmented regions, instead of all texture edges.
 
To the best of our knowledge, this is the first attempt at facilitating cross-modality UDA for segmentation with the adaptation of self-semantic edge delineation. Specifically, we propose a novel cascaded multi-task framework with the consecutive semantic contour delineation and an edge map conditioned semantic segmentation network. The adaptation of the semantic contour delineation is achieved by an edge map adversarial alignment along with supervised delineation in the source domain. We note that the edge label in the source domain can be directly generated from its segmentation label with the Canny edge detector \cite{ding2001canny} in a self-supervision manner, and we do not need an additional labeling effort. At the testing stage, the segmentation network takes the concatenated MRI slice and its edge map as input for inference. We follow the conventional adversarial segmentation UDA to enforce the feature level alignment. In addition, our framework is orthogonal to unsupervised learning approaches; therefore the  widely used self-entropy minimization can be simply added on top of our framework. We validate our framework on a brain tumor segmentation task using multiple MRI modalities to demonstrate the effectiveness of our framework. 

\section{Methodology}  \vspace{-5pt} 
\subsection{Conventional Adversarial UDA for Segmentation}\vspace{-5pt} 
In the cross-modality segmentation UDA, we have a labeled source domain with the sample $\{x^s, y^s\}$ and an unlabeled target domain with the sample $\{x^t\}$. We note that the label $y^s$ is the pixel-wise segmentation map. We target to learn a mapping $f:x\rightarrow y$ that performs well in both source and target domains with the training data of $\{x^s, y^s\}$ and $\{x^t\}$.

A typical solution would be to train an autoencoder-based segmentation network with $\{x^s, y^s\}$, following a supervise learning scheme, and simultaneously to train a feature-level discriminator, following an adversarial game \cite{liu2021dual}. Specifically, we use a cross-entropy loss in each pixel for source domain supervised learning, given by
\begin{align}
\mathcal{L}^y_{CE} = -\sum_i^C {y}_i^s\text{log}(\tilde{y}_i^s),
\end{align}where $\tilde{y}^s$ is the prediction of the segmentation network, and $C$ indicates the number of semantic categories. 

The discriminator takes the latent representation, i.e., the output features of the encoder $Enc(x^s)$ or $Enc(x^t)$, as input, and classify whether they are from the source or target domain.~The encoder tries to confuse the discriminator, by making $Enc(x^s)$ and $Enc(x^t)$ indistinguishable.~Then, both $Enc(x^s)$ and $Enc(x^t)$ can share the same decoder for segmentation. The to-be minimized binary cross-entropy loss of the feature-level discriminator can be formulated as:
\begin{align}
\mathcal{L}^f_{Dis} &=\mathbb{E}_{x^s} [\text{log}(Dis(Enc(x^s)))] \\&+ \mathbb{E}_{x^t} [\text{log}(1-Dis(Enc(x^t)))].
\end{align} 
Here, the encoder tries to maximize $\mathcal{L}^f_{Dis}$, i.e., minimizing $-\mathcal{L}^f_{Dis}$ in implementation \cite{salimans2016improved}. 

\subsection{Semantic Contour Guided Segmentation UDA}\vspace{-5pt} 

Our proposed work builds upon the adversarial UDA approach in the following way.~Instead of directly aligning the semantic segmentation features that have a large domain gap, we propose to achieve semantic contour delineation adaptation as a preparatory task. Since the low-level edge inherits a smaller domain gap, the delineation task itself can serve as a precursor module for the semantic segmentation \cite{yuheng2017image,rampun2019breast}. The overall framework is shown in Fig.~\ref{fig:illus}.

Specifically, we have an independent semantic contour network to predict the edge map of both source and target domain samples. For source domain samples, the ground truth semantic edge label $e^s$ can be simply generated from their segmentation label with the Canny edge detector \cite{ding2001canny}. The edge label has binary value, and the output layer uses a sigmoid unit. The corresponding CE loss in each pixel for the edge delineation can be expressed as: 
\begin{align}
\mathcal{L}^e_{CE} = -e^s\text{log}(\tilde{e}^s) - (1-e^s)\text{log}(1-\tilde{e}^s).
\end{align}

To achieve the adaptation of the semantic contour delineation, we simply adopt the edge map level adversarial training. The discriminator takes the predicted edge maps $\tilde{e}^s$ and $\tilde{e}^t$ as input and classify their domains.~The cross-entropy loss of the edge map discriminator can be formulated as:
\begin{align}
\mathcal{L}^e_{Dis} =\mathbb{E}_{\tilde{e}^s} [\text{log}(Dis(\tilde{e}^s))]+ \mathbb{E}_{\tilde{e}^t}[\text{log}(1-Dis(\tilde{e}^t))].
\end{align} 
The discriminator is trained by minimizing $\mathcal{L}^e_{Dis}$, while the semantic contour network tries to maximize it.

Next, we concatenate the semantic edge map with the original MRI slice, which is then input to the segmentation network. We note that the semantic edge map provides location information, which is consistent with the segmentation boundaries. In addition, our latent space encoding can be expressed as $Enc(x^s,\tilde{e}^s)$ and $Enc(x^t,\tilde{e}^t)$. Therefore, the adversarial loss in Eq. (2) can be reformulated as:
\begin{align}
\mathcal{L}^f_{Dis} &= \mathbb{E}_{x^s,\tilde{e}^s} [\text{log}(Dis(Enc(x^s,\tilde{e}^s)))]\\ &+ \mathbb{E}_{x^t,\tilde{e}^t}[\text{log}(1-Dis(Enc(x^t,\tilde{e}^t)))].
\end{align}

\begin{figure}[t]
\begin{center}
\includegraphics[width=1\linewidth]{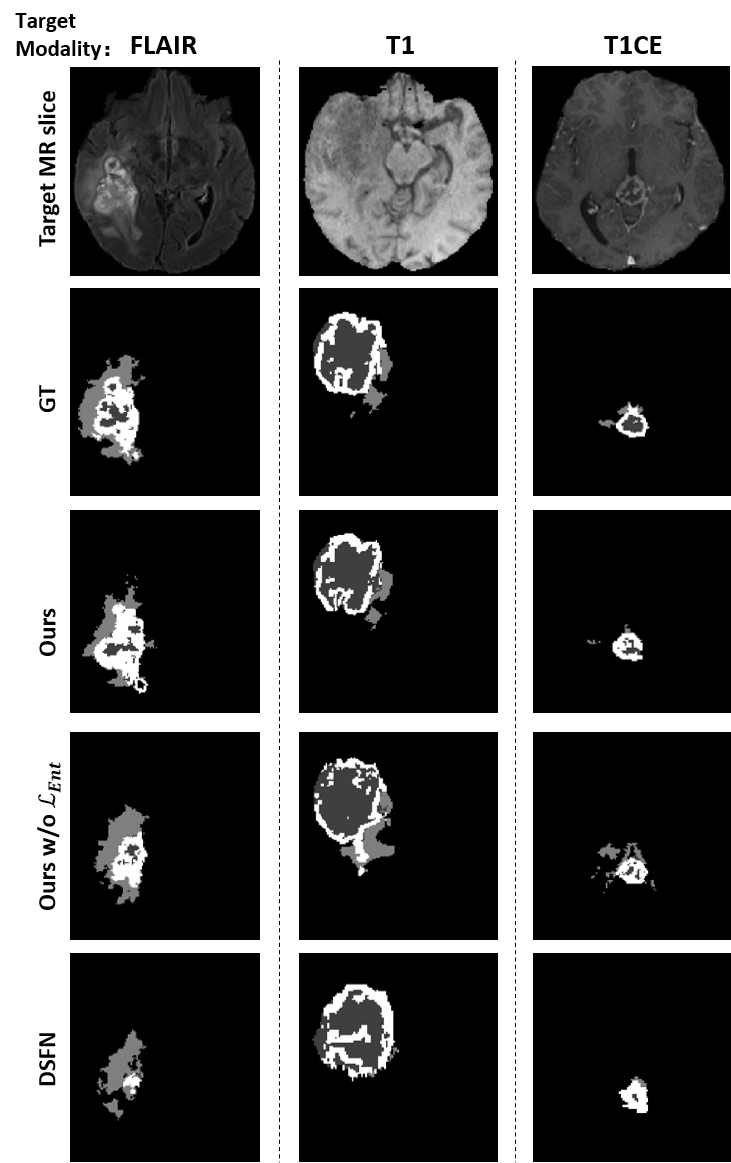}\vspace{-20pt}
\end{center}
\caption{Comparison with an UDA method, DSFN~\cite{zou2020unsupervised}, and an ablation study of the self-entropy minimization for adapting T2-weighted MRI to FLAIR/T1/T1CE MRI.} 
\label{exp1}
\end{figure} 

UDA to learn high-level semantic alignment in the unlabeled target modality can be seen as an unsupervised learning problem.~Therefore, unsupervised learning approaches can be applied to further improve the performance \cite{wang2020fully,bateson2020source}. To this end, we adopt self-entropy minimization, which can be easily incorporated into our framework. Self-entropy minimization encourages a confident output,~i.e., the maximum softmax value can be high~\cite{grandvalet2005semi,liu2022spie}. For the segmentation task, the self-entropy minimization can be formulated as the pixel-wise averaged entropy of the last layer's softmax prediction:
\begin{align}
    \mathcal{L}_{Ent}=
    \frac{1}{N} \sum_n^{N}\{{\sigma_{n} \text{log} \sigma_{n}}\},
\end{align}
where $N$ indicates the number of pixels of the segmentation map, and $\sigma_{b,n}$ is the histogram distribution of the softmax output of the $n$-th pixel. By minimizing the pixel-wise averaged entropy, the outputs are encouraged to approach a one-hot distribution.

In summary, we jointly minimize the following objectives for different modules:
\begin{align}
    &\text{Semantic Contour Net}:  \mathcal{L}^e_{CE} - \alpha\mathcal{L}^e_{Dis},\\
    &\text{Edge discriminator}: \mathcal{L}^e_{Dis},\\
    &\text{Encoder}: \mathcal{L}^y_{CE}-\beta\mathcal{L}^f_{Dis}+\lambda\mathcal{L}_{Ent},\\ &\text{Decoder}: \mathcal{L}^y_{CE}+\lambda\mathcal{L}_{Ent},\\
    &\text{Segmentation feature discriminator}: \mathcal{L}^f_{Dis},
\end{align}where $\beta$ and $\lambda$ are the weighting parameters.

At the testing stage, we first generate the semantic edge map of the input MRI slice with our semantic contour network. The segmentation network then makes the inference based on the concatenated edge map and its original MRI slice. Of note, we do not need discriminators in testing.

\section{EXPERIMENTS AND RESULTS} \vspace{-5pt}

\subsection{Data and Evaluation Protocol}  \vspace{-5pt}

We evaluated our framework on the publicly available multi-modality BraTS2018 database \cite{menze2014multimodal}. The database contains a total of 285 subjects, in which 210 subjects have high-grade gliomas, and the remaining 75 subjects have low-grade gliomas. Each subject has four modalities that are aligned, including T1-weighted (T1), T1-contrast enhanced (T1ce), T2-weighted (T2), and T2 Fluid Attenuated Inversion Recovery (FLAIR) MRI. The label has three classes for each pixel, including EnhT, ED, and CoreT \cite{menze2014multimodal}. In addition, the sum of EnhT, ED, and CoreT represents the whole tumor \cite{liu2021adapting}.

In order to demonstrate the effectiveness of our framework, we followed the standard cross-modality UDA setting that selects 80\% subjects for training and 20\% subjects for testing  \cite{han2021deep,zou2020unsupervised}.~We used T2-weighted MRI as the labeled source domain, and FLAIR, T1-weighted, and T1CE MRI as the unlabeled target domains \cite{han2021deep,zou2020unsupervised}.~We normalized the image intensity to [-1, 1] and applied cropping and rotation for data augmentation as in \cite{han2021deep,zou2020unsupervised}. We note that the data were used in an unpaired manner \cite{zou2020unsupervised}, and we did not need additional edge labels.
 
We adopted the Dice score and Hausdorff distance for quantification as in \cite{zou2020unsupervised}. The Dice score measures the overlaps between ground truth labels and predictions, and the larger value indicates better performance. In contrast, the Hausdorff distance is defined between two sets of points in the metric space, and the lower value indicates the better performance.

\begin{table}[t]
\centering
\caption{Comparison of Core/EnhT/Whole tumor segmentation for the cross-modality UDA. Results are averaged over T2-weighted to T1-weighted, T1CE, and FLAIR MRI. We used the same segmentation backbone as in DSFN~\cite{zou2020unsupervised}.} 
\resizebox{1\linewidth}{!}{
\begin{tabular}{c|ccc|ccc}
\hline

& DICE & Score & [$\%$] $\uparrow$  & Hausdorff & Distance & [mm] $\downarrow$ \\ \cline{2-7}

{Method}& \textbf{CoreT} & \textbf{EnhT} & \textbf{Whole}  & \textbf{CoreT} & \textbf{EnhT} & \textbf{Whole}  \\ \hline \hline

No UDA~\cite{zou2020unsupervised} &20.4 &39.7 &22.6  &55.4 &67.3 &46.8 \\\hline  
    
CycleGAN~\cite{zhu2017unpaired}  &52.6& 41.9& 57.8 &35.3 &48.2 &21.9\\ 

SIFA~\cite{chen2019synergistic} &54.4 &41.1& 59.3  &28.6& 35.7& 17.1\\

DSFN~\cite{zou2020unsupervised} &56.8 &42.7 &66.1& 27.8 &34.5& 15.6\\  \hline

Ours w/o $\mathcal{L}_{Ent}$ &57.6 &43.9 &67.4& 26.6 &  33.5 &13.9 \\

Ours  &\textbf{58.0} &\textbf{44.6}  &\textbf{67.9} &\textbf{26.4}  & \textbf{33.2}  &\textbf{13.5} \\\hline
 
\end{tabular}}\label{tab1}\vspace{-8pt}  
\end{table}

\subsection{Network and Training Details} \vspace{-5pt} 

In order to demonstrate the generality of our framework, we adopted two backbones for our segmentation network as in \cite{zou2020unsupervised, han2021deep}. In \cite{zou2020unsupervised}, the encoder and decoder based segmentation network consists of the residual blocks and dilated residual blocks, respectively.~The segmentation network used in \cite{han2021deep} is based on the more powerful  Deeplab-ResNet50.~We followed~\cite{yang2016object} to use a fully convolutional encoder-decoder network for our semantic contour network, and only revised the input layer to a single channel for our MRI data.~Our segmentation feature discriminator is composed of 3 convolutional and 2 fully connected layers.~The edge map discriminator has 4 convolutional and 2 fully connected layers.  
 
We implemented our framework using the PyTorch toolbox.~The training was implemented in a server with an NVIDIA V100 GPU, which took about 5 hours.~In testing, the inference took only 0.1s. The learning rates were set as $1\mathrm{e}{-3}$ and $1\mathrm{e}{-4}$ for segmentation and semantic contour networks and discriminators, respectively. We used the momentum of 0.5 consistently.

\subsection{Qualitative Evaluations} \vspace{-5pt} 

Fig.~\ref{exp1} shows a visualization of the segmentation maps of the samples from different target domains. We used the segmentation network as in DSFN~\cite{zou2020unsupervised} for a fair comparison. Considering the large appearance discrepancies between different MRI modalities, the performance dropped significantly. Even with the adaptation via DSFN~\cite{zou2020unsupervised}, the results were significantly different from the ground truth as visually assessed, potentially leading to severe misdiagnosis. By contrast, we can see that our proposed framework yielded superior segmentation results, compared with DSFN~\cite{zou2020unsupervised}, where our framework is based on a relatively simple dual-scheme fusion network. Therefore, our self-semantic contour network is considered a simple yet efficient add-on module to guide the challenging semantic segmentation adaptation. In addition, the self-entropy minimization was able to further boost the performance in an add-on fashion.

\begin{table}[t]
\centering
\caption{Whole tumor segmentation performance of the cross-modality UDA. We used the same segmentation backbone as in DSA~\cite{han2021deep}.} 
\resizebox{1\linewidth}{!}{
\begin{tabular}{c|ccc|ccc}
\hline

& DICE & Score & [$\%$] $\uparrow$  & Hausdorff & Distance & [mm] $\downarrow$ \\ \cline{2-7}

{Method} & \textbf{FLAIR} & \textbf{T1} & \textbf{T1CE} & \textbf{FLAIR} & \textbf{T1}  & \textbf{T1CE}  \\ \hline \hline

No UDA~\cite{han2021deep}  &65.1 &4.2&6.3   &28.0 &55.7&49.8 \\\hline  
    


 
DSA~\cite{han2021deep}  &81.8 &57.7&62.0 &8.6 &14.2 &13.7 \\\hline

Ours w/o $\mathcal{L}_{Ent}$ &82.6&58.8 &63.1 &8.1  &12.8 &11.5 \\

Ours  &\textbf{82.9}  &\textbf{59.3} &\textbf{63.5}  & \textbf{7.9} &\textbf{12.5}  &\textbf{11.2}  \\\hline




\end{tabular}}\label{tab2}\vspace{-8pt}  
\end{table}

\subsection{Quantitative Evaluations}\vspace{-5pt} 

Tables~\ref{tab1} and~\ref{tab2} show the quantitative evaluation results of the segmentation network backbone using DSFN~\cite{zou2020unsupervised} and DSA~\cite{han2021deep}, respectively. Our proposed framework yielded superior performance, compared with the competing methods with the same segmentation backbone, thus demonstrating its effectiveness and generality. The ablation study of the self-entropy minimization was also consistent with the qualitative evaluations. Since we used T2-weighted MRI as our source domain, there is a relatively smaller domain gap between T2-weighted and T2-FLAIR MRI, while the domain gap is larger between T2-weighted and T1-weighted or T1CE MRI. In Table \ref{tab2}, accordingly, we can see that the performance in T1-weighted or T1CE MRI domains is significantly inferior to FLAIR MRI.  

\section{Discussion and Conclusion} \vspace{-5pt}  

In this work, we proposed a multi-task adaptation framework to utilize the low-level edge detection task to facilitate the challenging semantic segmentation.~The semantic contour has been a precursor task for semantic segmentation, since the edge information is highly transferable, thus simplifying the adaptation of the semantic delineation.~With the relatively precise contour in the target domain, we were able to guide the subsequent segmentation, by providing the boundaries of semantic classes.~We did not need an additional edge label, but instead, we resorted to the simple Canny operator to acquire the source domain edge map in a self-supervision manner. In addition, the widely used self-entropy minimization was incorporated to boost the performance. We demonstrated its effectiveness on the T2-weighted to FLAIR/T1-weighted/T1CE MRI adaptation task and achieved superior performance, compared with the competing state-of-the-art methods.   

\section{COMPLIANCE WITH ETHICAL STANDARDS} \vspace{-8pt}
This research study was conducted retrospectively using human subject data made available in open access by
\href{https://www.med.upenn.edu/sbia/brats2018/data.html}{BraTS'18}. 

\section{ACKNOWLEDGMENTS}\vspace{-8pt}
{This work is partially supported by NIH R01DC018511, R01DE027989, and P41EB022544.}

\bibliographystyle{IEEEbib}

\bibliography{ref5page}

\end{document}